\documentclass{article}
\usepackage{spconf,amsmath,graphicx}

\usepackage{microtype}
\usepackage{graphicx}
\usepackage{subcaption}
\usepackage{booktabs} 

\usepackage{algorithm}               
\usepackage{algorithmic}             

\usepackage{color}
\usepackage{hyperref}
\usepackage{amsmath}
\usepackage{amsthm}
\usepackage{amssymb}

\usepackage{multirow}
\usepackage{diagbox}

\newtheorem{definition}{Definition}[]
\usepackage{enumitem}

\title{Non-Recursive Graph Convolutional Networks}
\name{Hao Chen$^{1,2}$, Zengde Deng$^{3}$, Yue Xu$^{4}$, Zhoujun Li$^{1}$}
\address{$^1$State Key Lab of Software Development Environment, Beihang University, Beijing, China. \\$^2$Tencent Inc., Shenzhen, China. $^3$Cainiao Network, Hangzhou, China. $^4$Alibaba Group, Hangzhou, China.}
\begin{document}
%
\maketitle
\begin{abstract}
Graph Convolutional Networks (GCNs) are powerful models for node representation learning tasks. 
However, the node representation in existing GCN models is usually generated by performing recursive neighborhood aggregation across multiple graph convolutional layers with certain sampling methods, which may lead to redundant feature mixing, needless information loss, and extensive computations.
Therefore, in this paper, we propose a novel architecture named Non-Recursive Graph Convolutional Network (NRGCN) to improve both the training efficiency and the learning performance of GCNs in the context of node classification.
Specifically, NRGCN proposes to represent different hops of neighbors for each node based on inner-layer aggregation and layer-independent sampling.
In this way, each node can be directly represented by concatenating the information extracted independently from each hop of its neighbors thereby avoiding the recursive neighborhood expansion across layers.
Moreover, the layer-independent sampling and aggregation can be precomputed before the model training, thus the training process can be accelerated considerably.  
Extensive experiments on benchmark datasets verify that our NRGCN outperforms the state-of-the-art GCN models, in terms of the node classification performance and reliability. 
\end{abstract}
\vspace{-1em}
\section{Introduction}
\begin{figure*}[t]
	\centering
	\includegraphics[width=0.95\textwidth]{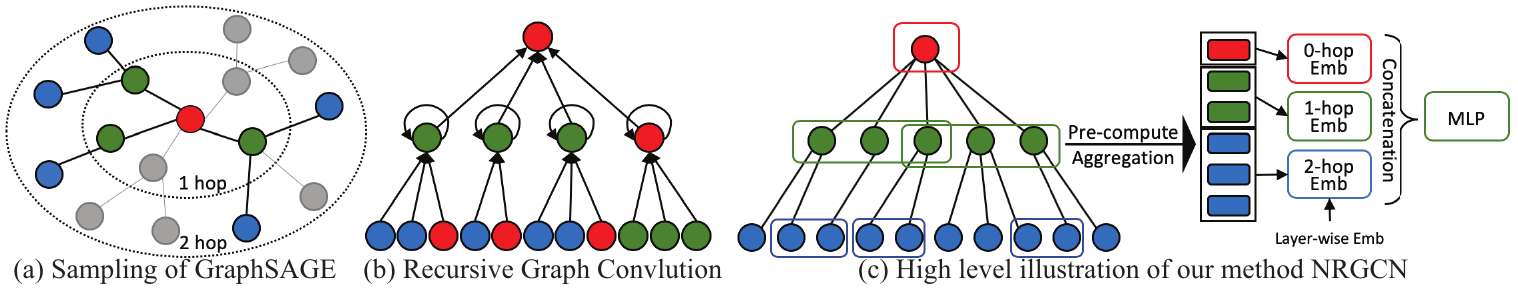}
	\vskip -1.0em
	\caption{Illustrations of recursive GCNs and NRGCN. The red, green and blue nodes denote the root node, the $1$-hop neighbors and $2$-hop neighbors, respectively. a) The sampling process of GraphSAGE, where the grey nodes denote the discarded neighbors. b) The message passing path of the recursive GCNs. c) Our proposed NRGCN. On the left, we present the layer-wise multiple sampling, where $P_1=1,S_1=2,L_1=3;P_2=2,S_2=3,L_2=2$. On the right, we demonstrate the inner-layer aggregation and the concatenation of the layer-wise embeddings.}
	\label{fig:framework}
	\vspace{-2em}
\end{figure*}

\vspace{-1em}
Convolutional Neural Networks (CNNs)~\cite{lecun1995cnn,lecun1998cnn,he2016resnet}, have revolutionized various machine learning tasks on grid-like data such as image classification and generation. In recent years, motivated by CNNs, Graph Convolutional Networks (GCNs)~\cite{bruna2013spectral} have been proposed to learn from the graph-structured data by stacking multiple first-order graph convolution layers~\cite{kipf2016gcn,dropedge,sgc,velivckovic2017gat}. However, the computational complexity of GCN models grows exponentially with the number of GCN layers due to the recursive neighborhood aggregation across consecutive layers, which can be prohibitive for large-scale datasets. 
Therefore, recent works proposed various sampling methods to accelerate the convolution process and reduce the training burden by discarding part of the neighbors for each node in a graph~\cite{gunets,lagcn,chen2018fastgcn,chiang2019cluster,hamilton2017graphsage,huang2018asgcn,graphsaint}. For example, Figure~\ref{fig:framework}~(a) presents a typical layer-dependent sampling strategy adopted by GraphSAGE~\cite{hamilton2017graphsage}.

However, the recursive aggregation based on current sampling strategies encounters the following issues. 
First, the aggregated information could be redundant. For example, as shown in Figure \ref{fig:framework} (b), during the bottom-up recursive graph convolution, the information of the root~(red) nodes and the intermediate~(green) nodes have been aggregated and updated for several times, which may lead to chaotic feature mixing and redundant computations. 
Second, the layer-dependent sampling strategy, such as GraphSAGE~\cite{hamilton2017graphsage}, may lose important information since some nodes which contain crucial content could be discarded. Consequently, this could disturb the final predictions. 
On the other hand, recent work also proposed layer-wise sampling methods such as FastGCN~\cite{chen2018fastgcn} and ASGCN~\cite{huang2018asgcn}. However, according to~\cite{graphsaint}, the sampling of FastGCN becomes too sparse to achieve high accuracy while ASGCN imports expensive sampling algorithm and extra sampler parameters to be learned.

In this paper, we propose a novel graph convolution architecture named Non-Recursive Graph Convolutional Network (NRGCN) which introduces a more principled layer-independent multi-sampling strategy along with a tailored inner-layer aggregation method to improve both the training efficiency and the learning performance of GCNs in the context of node classification.
Specifically, as shown in Figure~\ref{fig:framework} (c), for each node, NRGCN performs layer-wise multi-sampling at each layer by repeating the following two steps: first, NRGCN samples a subset of parent nodes at the previous layer and then union all the neighbors of the sampled parent nodes as a localized node set; next, NRGCN samples a given number of nodes from the localized node set as one sampled node set at the current layer. In this way, NRGCN is able to perform layer-independent sampling to avoid recursive neighborhood aggregation across layers while still preserving the localized graph structure. 
Moreover, as shown in Figure~\ref{fig:framework} (c), NRGCN proposes an inner-layer aggregation method to represent each hop of neighbors with the multi-sampling subsets and concatenates the layer-wise embeddings to represent the root node.
Specifically, during the inner-layer aggregation, the aggregation of the sampled node subsets can be precomputed before the model training as a prepossessing step to largely reduce the training complexity. 
Finally, we conduct extensive experiments on benchmark node classification datasets to verify the effectiveness of NRGCN compared with the state-of-the-art GCN models. 


To sum up, the main contribution is three-fold:
1)	We propose the layer-independent sampling strategy for NRGCN which can avoid recursive neighborhood aggregation across layers while still preserving the localized graph structure. 
2) 	We propose the inner-layer aggregation method through which the node representations in NRGCN can be precomputed before the model training as a prepossessing step so as to largely reduce the training complexity. 
3) 	We conduct extensive experiments on benchmark datasets to verify that the proposed NRGCN outperforms the state-of-the-art GCN models and shows superior robustness under graph adversarial attacks, in terms of the node classification performance.

\vspace{-1em}
\section{Proposed Method}
\vspace{-1em}

\subsection{Preliminaries}
\vspace{-1em}
\smallskip\noindent\textbf{Notations.} This paper mainly focuses on the graph-structure datasets $\mathcal{G}=(\textbf{X}, \textbf{A})$. $\textbf{X} = \{ \textbf{x}_1, \textbf{x}_2, \cdots, \textbf{x}_N \}$ denotes the node features of the nodes $\textbf{V} = \{v_1, \cdots, v_N\}$. 
The $(i,j)$-th entry of adjacency matrix $\textbf{A} \in \mathbb{R}^{N\times N}$ represents the weight associated with edge $(v_i, v_j)$. Note that the adjacency matrices of the graphs discussed throughout our paper are all binary. $\mathcal{N}_v$ denotes the set of $1$-hop neighbors of node $v$. $\mathcal{N}_v^l$ denotes the set of $l$-hop neighbors of $v$.

\smallskip\noindent\textbf{Structure of GCNs.} The core structure of GCNs is the recursively stacked first-order convolutional layer, which can be viewed as a sequence of local smoothing processes \cite{li2018laplacian, taubin1995laplacian}. 
A single convolutional layer aggregates the hidden embedding generated by the previous smoothing process and outputs an embedding via a nonlinear function, which is given as
\begin{equation}
	\setlength{\abovedisplayskip}{3pt} 
	\setlength{\belowdisplayskip}{2pt}
	\textbf{h}_{v}^{(l+1)} = \sigma(\textbf{FC}_{\theta_{(l)}}(\sum_{u\in{\mathcal{N}_v}} w^{(l)}_{u,v} \textbf{h}^{(l)}_u )),
	\label{equ:gcn}
\end{equation}
where $\textbf{FC}_{\theta_{(l)}}$ refers to the fully-connected layer at layer $l$ with the parameter $ \theta_{(l)} $, $w^{(l)}_{v,u}$ denotes the aggregation weight of the neighboring nodes; $\sigma(\cdot)$ denotes the nonlinear activation function. Finally, GCNs predict the labels of nodes by feeding the hidden embeddings of the last layer $K$ to a MLP network $f(\cdot)$, \emph{e.g.}, $\hat{Y}_v = f(h_v^{(K)})$.
\vspace{-1em}
\subsection{Framework of NRGCN} 
\vspace{-0.5em}
As illustrated in JKNET~\cite{jknet}, the performance of current GCNs can be enhanced by connecting the hidden embeddings in the output layer as
\begin{equation}
	\setlength{\abovedisplayskip}{3pt} 
	\setlength{\belowdisplayskip}{2pt}
	\hat{Y}_v = f(\textbf{h}_v^{(0)}\|\textbf{h}_v^{(1)}\|\cdots\|\textbf{h}_v^{(K)}),
	\label{equ:jknet}
\end{equation}
where $\|$ denotes the concatenation operation.
Concretely, $\textbf{h}_v^{(l)}$ contains the information from the central node to the $l$-hops neighbors, which means, the information of closed neighbors has been mixed and recomputed several times. On the opposite, NRGCN utilizes a neat way to extract and blend the information of each hop of neighbors. Here we use $\hat{\textbf{h}}_v^{(l)}$ to denote the information that directly extracted from $l$-hop neighbors.
\begin{equation}
	\setlength{\abovedisplayskip}{3pt} 
	\setlength{\belowdisplayskip}{2pt}
	\hat{Y}_v = f(\hat{\textbf{h}}_v^{(0)}\|\hat{\textbf{h}}_v^{(1)}\|\cdots\|\hat{\textbf{h}}_v^{(K)}).
	\label{equ:nrgcn}
\end{equation}
By separating the embedding computation of different hop of neighbors, the embedding of different layers can be pre-computed and pre-saved independently to accelerate the training process. In the following sections, we introduce how to extract the $\textbf{h}_v^{(l)}$ from any given hop of neighbors while preserving the graph structure.
\vspace{-1em}
\subsection{Layer-Independent Multi-Sampling}
\vspace{-0.5em}
Firstly, it has been verified that GCNs benefit from the aggregation of the localized neighbors which preserves the graph structure and the node features~\cite{li2018laplacian}. Thus, the main challenge of the layer-wise sampling is how we restrict the sampling nodes to localized places, besides, this sampling method should be able to be adapted to any given hop of neighbors. Secondly, the current sampling-based GCNs select neighbors based on the sampled nodes at the previous layer, which is layer-dependent. To address this issue, the sampling strategy of NRGCN aims at getting rid of the dependency across layers and enable a multi-sampling approach to decrease information loss.

\smallskip\noindent\textbf{Layer-wise sampling.} The layer-wise sampling is a two-step process. We first sample the parentt node set and then union the neighbor set of the selected parents to build the localized node set. Then we sample specify amounts of nodes out of the localized node set. The formal definition of localized node set can be given as follows.
\vspace{-0.5em}
\begin{definition}{\rm (Localized node set)}
	Given $\mathcal{P}$ as the parent set, the localized node set can be defined as
	\begin{equation}
		\setlength{\abovedisplayskip}{3pt} 
		\setlength{\belowdisplayskip}{2pt}
		\hat{\mathcal{N}}(\mathcal{P}) = \bigcup_p \hat{\mathcal{N}}_p, \ \forall p \in \mathcal{P}.
	\end{equation}
\end{definition}
\vspace{-1em}
Based on the above definition, now we start from the sampling of the 1-hop neighbors and present how to sample the parent set $\mathcal{P}$. For the sampling of 1-hop neighbors, the parent set $\mathcal{P}=\{v\}$ is set as the root node itself, where $\hat{\mathcal{N}}(\mathcal{P}) = \mathcal{N}_v$. When sampling the neighbors of $l$-hop neighbors, we sample the parent set from the $(l-1)$-hop neighbors~(i.e., $\mathcal{P} \subset \mathcal{N}_v^{l-1}$). Then we collect the neighbors of the parent nodes set $\mathcal{P}$ and construct the localized node set $\hat{\mathcal{N}}(\mathcal{P})$. After that, NRGCN samples $L_l$ nodes out of $\hat{\mathcal{N}}(\mathcal{P})$, where $L_l$ denotes the sample size for $l$-hop neighbors. 

\smallskip\noindent\textbf{Multiple Sampling.}
As shown in Figure~\ref{fig:framework}, GraphSAGE random selects several 1-hop neighbors and recursively samples neighbors hop by hop. As a consequence, GraphSAGE may drop valuable 1-hop neighbors. To remit the information loss, NRGCN uses multiple sampling to better extract the information of any given hop of neighbors. 

Take $S^l$ as the sampling times for $l$-hop neighbors, we repeat the sampling process $S_l$ times to obtain the sampling set $\textbf{S}_v^l = \{\textbf{s}^{(l,1)}_v, \cdots, \textbf{s}^{(l,S_l)}_v\}$.
Given the total layers of NRGCN as $K$, for hops from $0$ to $K$ on node $v$, we can get the sampling set as $\textbf{S}_v = \{\textbf{s}^{(0,1)}_v,\textbf{s}^{(1,1)}_v, \cdots, \textbf{s}^{(l,S_1)}_v,\cdots, \textbf{s}^{(K,1)}_v, \cdots, \textbf{s}^{(K,S_K)}_v\}$, where $\textbf{s}^{(0,1)}_v = \{v\}$.

\vspace{-1em}
\subsection{Inner-Layer Aggregation.} 
\vspace{-0.5em}
Currently, GCNs are suffering from the computational burden caused by recomputing the convolution process during each inference process in both training and testing. As NRGCN using a layer-independent sampling method, the embedding of each layer can be computed independently. Thus, NRGCN can pre-compute and pre-save the embedding of each hop of neighbors to acclerate the training process.

Different from the common aggregation process which collects the embeddings generated by nonlinear function, NRGCN directly aggregates raw features of sampled nodes in a parameter-free manner, which can be pre-computed and stored in memories. This parameter-free aggregation process can be written as
\begin{equation}
	\setlength{\abovedisplayskip}{3pt} 
	\setlength{\belowdisplayskip}{2pt}
	\textbf{e}^{l,i}_v = \textbf{Agg}\{\textbf{x}_u, \forall u \in \textbf{s}^{(l,S_l)}_v \},
	\label{equ:setagg}
\end{equation}
where $\textbf{Agg}(\cdot)$ denotes the aggregation function. Note that we only consider the mean aggregator \footnote{Other aggregators can also be easily adapted to our model \cite{hamilton2017graphsage,zhang2018gaan}} in this paper. 
Thus, for any root node $v$, an embedding matrix associated with $v$ is given as 
\begin{equation}
	\setlength{\abovedisplayskip}{3pt} 
	\setlength{\belowdisplayskip}{2pt}
	\textbf{E}_v = \{e_v^{(0,1)}, e_v^{(1,1)},\cdots, e_v^{(1,S^1)}, \cdots, e_v^{(K,S^1)}, \cdots, e_v^{(K,S^K)}\},
\end{equation}
where $\textbf{E}_v \in \mathbb{R}^{(\sum_{l=0}^K S^l) \times F}$ and $S_0 = 1$. The embeddings of all the nodes can be stored in a tensor $\mathcal{E} = \{\textbf{E}_1,\cdots,\textbf{E}_N\} \in \mathbb{R}^{N\times(\sum_{l=0}^K S^l) \times F}$. In~\textsection~\ref{sec:result}, we demonstrate that NRGCN can still achieve comparable performance when all the $S_l$ are set as 1. Thus, the size of the embedding tensor shrinks to $N\times K \times F$, which can be easily stored even for commercial large-scale graphs.

In order to get the hop embeddings $\hat{\textbf{h}}_v^{(l)}$ in Equ \ref{equ:nrgcn} with low model complexity, NRGCN use a shared nonlinear fully-connected layer to map the node embeddings $e_v^{l,i}$ to the same hidden space. Then the hidden embeddings of the same hop are aggregated as
\begin{equation}
	\setlength{\abovedisplayskip}{3pt} 
	\setlength{\belowdisplayskip}{2pt}
	\hat{\textbf{h}}^{(l)}_v = \textbf{Agg} \{
	\sigma(\textbf{FC}_\theta (\textbf{e}_v^{l,i})), {\rm \ for} \ i =1,\cdots ,S^l\},
	\label{equ:hopagg}
\end{equation}
where $\textbf{FC}_\theta(\cdot)$ denotes the transition function which is shared by all the hops and $\sigma(\cdot)$ denotes the nonlinear function.

\vspace{-1em}
\section{Experiments}
\vspace{-1em}
In this section, we first evaluate our model on four public node classification datasets. 
Next, we investigate the reliability of NRGCN on two adversarial attack datasets. At last, we compare the training complexity and analyze the parameter sensitivity.
\vspace{-1em}
\subsection{Experiment Setup} 
\vspace{-0.5em}
We compare NRGCN with other baselines on four public benchmarks: Cora, Citeseer, Pubmed, and Reddit \cite{hamilton2017graphsage}. We split the datasets into training, validation, and testing sets in a commonly used way as in previous works~\cite{chen2018fastgcn,hamilton2017graphsage,huang2018asgcn}. We compare the performance of NRGCN with six baselines: GCN~\cite{kipf2016gcn}, GAT~\cite{velivckovic2017gat}, JKNET~\cite{jknet}, GraphSAGE~\cite{hamilton2017graphsage}, FastGCN~\cite{chen2018fastgcn}, ASGCN~\cite{huang2018asgcn}. For GCN, GAT, GraphSAGE, FastGCN, and ASGCN, we set the number of hidden layers as two. For JKNet, we report the best scores when choosing the hidden layers from 2, 4, 8, 16, 64~\cite{dropedge}. For sampling-based methods, such as GraphSAGE, we set the sampling number of layer 1 as $S_g^1 = 5$ for citation dataset and as $S_g^1 = 15$ for Reddit. The sampling number of layer 2 is set as $S_g^2 = 5$ for citation datasets and as $S_g^2 = 25$ for Reddit. 

Here we evaluate two variations of NRGCN: 1) NRGCN-SIN: the single sampling version of NRGCN, where the sampling parameters are set as $P_1=1, S_1=1, L_1=5, P_2=5, S_2=1, L_2=25$ for citation datasets and $P_1=1, S_1=1, L_1=15, P_2=15, S_2=1, L_2=375$ for Reddit. In our implementation, the sampling size and method of NRGCN-SIN is set as the same as GraphSAGE to make a fair compariation. 2) NRGCN-MUL: the multiple sampling version of NRGCN, where the sampling parameters are set as $P_1=1, S_1=5, L_1=3, P_2=2, S_2=5, L_2=5$ for citation datasets and $P_1=1, S_1=15, L_1=5, P_2=2, S_2=15, L_2=25$ for Reddit. The $f(\cdot)$ in Equation~\ref{equ:nrgcn} is set as a 2 layer MLP network.
The hidden dimension for the citation network datasets (i.e., Cora, Citeseer, and Pubmed) is set as 128. For Reddit, the hidden dimension is selected to be 256 as suggested by \cite{hamilton2017graphsage}. The batch size of the citation network is chosen to be 256 and 64 for Reddit.

\begin{table}[t]
	\small
	\centering
	\caption{Accuracy comparison with baseline methods.}
	\vspace{-1em}
	\setlength{\tabcolsep}{2.7mm}{
		\begin{tabular}{ c  c  c  c  c }
			\toprule
			Method & Cora & Citeseer & Pubmed & Reddit \\ \midrule
			GCN~\cite{kipf2016gcn} & 0.857 & 0.737 & 0.881 & --\\
			GAT~\cite{velivckovic2017gat} & 0.864 & 0.743 & 0.876 & --\\
			FastGCN~\cite{chen2018fastgcn} & 0.850 & 0.776 & 0.880 & 0.937\\
			JKNet~\cite{jknet} & 0.853 & 0.759 & 0.889 & 0.968 \\
			GraphSAGE~\cite{hamilton2017graphsage} & 0.862 & 0.783 & 0.881 & 0.943\\
			ASGCN~\cite{huang2018asgcn} & 0.874 & 0.796 & 0.906 & 0.963\\
			
			\midrule
			NRGCN-SIN & 0.872 & 0.794 & 0.904 & 0.967 \\
			NRGCN-MUL & \textbf{0.886}& \textbf{0.810} & \textbf{0.912}& \textbf{0.972}\\
			\bottomrule
	\end{tabular}}
	\label{tab:accu}
	\vspace{-0.75em}
\end{table}

\begin{table}
	
	\small
	\centering
	\caption{Accuracy comparison when under attacks.}
	\vspace{-1em}
	\setlength{\tabcolsep}{0.3mm}{
		\begin{tabular}{ccccccccc}
			\toprule
			
			\multicolumn{1}{l}{}        & \multicolumn{4}{c}{Cora}      & \multicolumn{4}{c}{Pubmed}     \\ \midrule
			\multicolumn{1}{c}{\footnotesize {Settings}} & \footnotesize {T} & \footnotesize {GraphSAGE} & \footnotesize{ASGCN} & \footnotesize{NRGCN}   & \footnotesize {T}  & \footnotesize {GraphSAGE} & \footnotesize{ASGCN} & \footnotesize{NRGCN}   \\\midrule
			\multirow{4}{*}{\footnotesize {Evasion}}  & 1 & 0.798      & 0.803 & \textbf{0.813} & 2  & 0.848   & 0.876 & \textbf{0.879} \\
			& 2 & 0.742    & \textbf{0.759} & 0.758 & 5  & 0.830     & 0.860 & \textbf{0.866} \\
			& 3 & 0.720     & 0.733 & \textbf{0.739} & 10 & 0.833     & 0.854 & \textbf{0.865} \\
			& 4 & 0.699     & 0.712 & \textbf{0.718} & 15 & 0.824     & 0.847 & \textbf{0.861} \\ \midrule
			\multirow{4}{*}{\footnotesize {Poisoning}} & 1 & 0.849    & 0.843 & \textbf{0.853} & 2  & 0.896     & 0.887 & \textbf{0.896} \\
			& 2 & 0.837     & 0.835 & \textbf{0.845} & 5  & 0.871     & 0.873 & \textbf{0.900} \\
			& 3 & 0.815     & 0.830 & \textbf{0.841} & 10 & 0.867     & 0.865 & \textbf{0.901} \\
			& 4 & 0.805     & 0.810 & \textbf{0.825} & 15 & 0.859     & 0.858 & \textbf{0.902} \\ \bottomrule 
	\end{tabular}}
	\label{tab:adversarial}
	\vspace{-1.5em}
\end{table}
\vspace{-1em}
\subsection{Result Analysis} 
\vspace{-0.5em}
\noindent\textbf{Performance of node classification.} 
\label{sec:result}
As shown in Table \ref{tab:accu}. NRGCN-SIN presents comparable or slightly better results than state-of-the-art baselines including GraphSAGE, while NRGCN-SIN is using the same sampling size and method as GraphSAGE. The comparable performance of NRGCN-SIN demonstrates that the layer-independent aggregation can extract same strong information as the recursive aggregation methods such as FastGCN, GraphSAGE, and ASGCN. Comparing NRGCN-SIN with JKNet, it demonstrates that the concatenation of layer-independent embeddings in NRGCN can result in a better performance than concatenating GCN hidden embeddings~(Equ~\ref{equ:jknet}). 
NRGCN-MUL outperforms other methods by a significant margin, showing the performance of NRGCN is benefitial from the multiple sampling approach.

\smallskip\noindent\textbf{Robustness under graph adversarial attacks.}
\label{sec:attacks}
Graph convolutional networks are shown to be vulnerable to adversarial attacks both at the testing process (evasion attacks) as well as the training process (poisoning attacks) \cite{zugner2019adversarial}. We evaluate our model on two attack scenarios: evasion attacks, where the graph neural networks are trained on the original graph and tested on the attacked graph; and poisoning attacks, where the graph neural networks are trained on the attacked graph and tested on the original graph. 

We utilize the basic DICE (i.e., `delete internally, connect externally')~\cite{zugner2019adversarial,dai2018adversarial,zugner2018adversarial,feng2019graph} to generate adversarial examples. 
In the evasion attacking evaluation, we randomly remove $T$ $1$-hop same-class neighbors for each node in the testing set. For the poisoning attacks, we remove $T$ $1$-hop same-class neighbors randomly for each node in the training set. We compare the performance of our method (NRGCN-MUL) with GraphSAGE \cite{hamilton2017graphsage} and ASGCN \cite{huang2018asgcn}. For Cora, we evaluate the performance by setting K ranging from 1 to 4. For Pubmed, we test $T$ ranging from 2 to 15. As shown in Table \ref{tab:adversarial} , when the graph structure is under attacks, NRGCN outperforms ASGCN and GraphSAGE. There are two main reasons: 1) the multiple layer-wise sampling can extract more information even the graph is under attack; 2) the concatenation of hop information can help the predict function $f(\cdot)$ to automatically choose reliable features, especially on Pubmed, where NRGCN is hardly affected by the poisoning attack.

\begin{figure}[t]
	\centering
	\includegraphics[width=1.0\columnwidth]{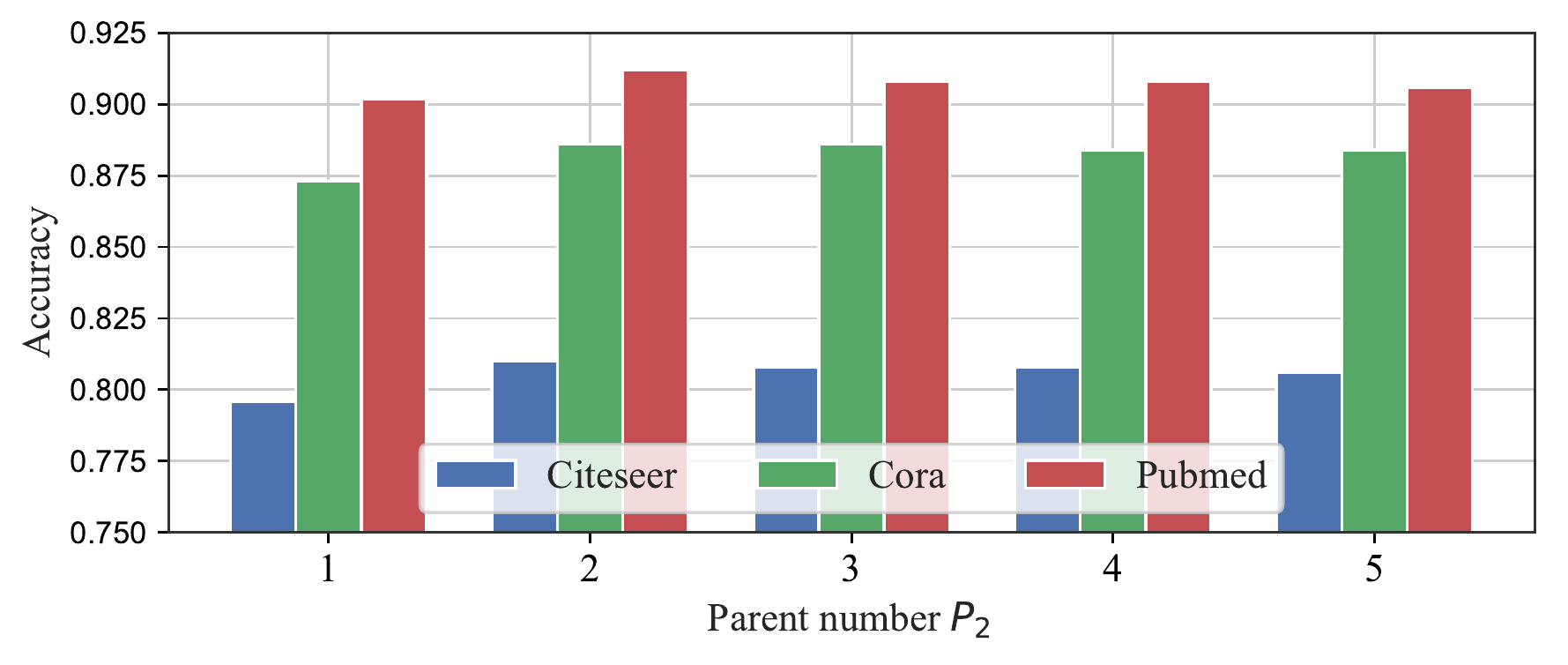}
	\vskip -1.0em
	\caption{Influence of different $P_2$.}
	\label{fig:abla}
	\vskip -1.5em
\end{figure}  

\smallskip\noindent\textbf{Complexity analysis.}
The computational cost of NRGCN mainly comes from a) aggregation process, b) training process. Here, for NRGCN, we denote the complexity of the feature aggregation for a given node as in Equation~\ref{equ:setagg} and~\ref{equ:hopagg} to be $\mathcal{O}_{NRagg} = \mathcal{O}(\sum_{l=0}^K S^lL^l)$. Relevantly, the complexity of the aggregation and mapping for recursive sampling method such as GraphSAGE is $\mathcal{O}_{Ragg} =\mathcal{O}(\prod_{l=0}^K S^l_g)$, where $S_g^l$ denotes the sampling number of layer $l$. $\mathcal{O}_{NRagg}$ and $\mathcal{O}_{Ragg}$ only differ with a constant coefficient which is less than 2~(under the settings of NRGCN-MUL), thus we denote both $\mathcal{O}_{NRagg}$ and $\mathcal{O}_{Ragg}$ as $\mathcal{O}_{agg}$. Similarly, we denote the complexity of the mapping function as $\mathcal{O}_{map}$, which also only differs with a constant coefficient. The number of total training epochs is denoted as $E$, the number of training nodes is donoted as $N_{train}$.
The recursive GCN models~(\emph{e.g.} GraphSAGE, ASGCN) perform recursive aggregations per training step, which arises a complexity of $E \cdot N_{train}\cdot \mathcal{O}_{agg} + E \cdot N_{train}\cdot \mathcal{O}_{map}$. 
Comparatively, NRGCN only performs Equation~\ref{equ:setagg} and~\ref{equ:hopagg} \textit{once} during the pre-processing step, which arises a complexity of $N \cdot \mathcal{O}_{agg} + E \cdot N_{train}\cdot \mathcal{O}_{map}$ in total. Note that $ N \cdot \mathcal{O}_{agg} \ll E \cdot N_{train}\cdot \mathcal{O}_{agg} $.

\smallskip\noindent\textbf{Parameter sensibility.}
In Figure~\ref{fig:abla}, we present the ablation study of the parent size of $2$-hop neighbors, which indicates that $P_2 >1$  could provide a consistent boost in accuracy compared with  $P_2 = 1$. Besides, $P_2=2$ presents the best performance on all datasets. Moreover, we find diminishing returns when sampling more node sets, indicating that we can balance the efficiency and performance to choose the proper sampling parameters.
\vspace{-1em}
\section{Conclusion}
\vspace{-1em}
    In this paper, we introduce the Non-Recursive Graph Convolutional Network(NRGCN) to alleviate the two issues existing in the recursive aggregation and "cut-off" sampling, which can capture more comprehensive information and aggregate the graph information without using recursive convolution. 
\textbf{Acknowledgement.}
This work is supported in part by the National Natural Science Foundation of China~(Grand Nos. U1636211, 61672081, 61370126) and 2020 Tencent Wechat Rhino-Bird Focused Research Program. We express our sincere gratitude to Dr. Dijun Luo for his constructive comments.
\newpage
\bibliographystyle{plain}
\bibliography{references}
\end{document}